\documentclass{article}

\usepackage[preprint]{neurips_2024}

\usepackage[utf8]{inputenc} 
\usepackage[T1]{fontenc}    
\usepackage{hyperref}       
\usepackage{url}            
\usepackage{booktabs}       
\usepackage{amsfonts}       
\usepackage{nicefrac}       
\usepackage{microtype}      
\usepackage{xcolor}         

\usepackage{amssymb}
\usepackage{amsmath}
\usepackage{booktabs}
\usepackage{xcolor}
\usepackage{tabularx}  
\usepackage{graphicx} 

\title{Mini Diffuser: Fast Multi-task Diffusion Policy
Training Using Two-level Mini-batches
}

\author{%
  Yutong Hu\thanks{yutong.hu@kuleuven.be}\\
  KU Leuven\\
  \And
  Pinhao Song\\
  KU Leuven\\
  \And
  Kehan Wen\\
  ETH Zurich\\
  \And
  Renaud Detry\\
  KU Leuven\\
}

\begin{document}

\maketitle
\thispagestyle{empty}
\pagestyle{empty}

\begin{abstract}

We present a method that reduces, by an order of magnitude, the time and memory needed to train multi-task vision-language robotic diffusion policies. This improvement arises from a previously underexplored distinction between action diffusion and the image diffusion techniques that inspired it: In image generation, the target is high-dimensional. By contrast, in action generation, the dimensionality of the target is comparatively small, and only the image condition is high-dimensional. Our approach, \emph{Mini Diffuser}, exploits this asymmetry by introducing \emph{two-level minibatching}, which pairs multiple noised action samples with each vision-language condition, instead of the conventional one-to-one sampling strategy. To support this batching scheme, we introduce architectural adaptations to the diffusion transformer that prevent information leakage across samples while maintaining full conditioning access. In RLBench simulations, Mini-Diffuser achieves 95\% of the performance of state-of-the-art multi-task diffusion policies, while using only 5\% of the training time and 7\% of the memory. Real-world experiments further validate that Mini-Diffuser preserves the key strengths of diffusion-based policies, including the ability to model multimodal action distributions and produce behavior conditioned on diverse perceptual inputs. Code available at: \href{https://mini-diffuse-actor.github.io/}{mini-diffuse-actor.github.io} along with videos and training logs.
\end{abstract}

\section{INTRODUCTION}

Diffusion models \cite{hoDenoisingDiffusionProbabilistic2020} have emerged as powerful generative tools due to their ability to model complex, multimodal distributions using iterative denoising processes. Initially popularized in image generation tasks \cite{rombachHighresolutionImageSynthesis2022}, diffusion models have recently demonstrated significant potential in decision-making areas like robotic control \cite{ajayConditionalGenerativeModeling2022, chiDiffusionPolicyVisuomotor2024, jannerPlanningDiffusionFlexible2022a, ze3DDiffusionPolicy2024, ke3DDiffuserActor2024}, showing competitive performance in both simulated benchmarks and real-world applications.

Despite their success, using diffusion models for action generation have a major limitation: they inherently require multiple denoising steps with condition-dependent predictions, leading to high computational costs during training and inference. Recent methods, such as DDIM \cite{songDenoisingDiffusionImplicit2020}, consistency models \cite{songConsistencyModels2023c}, and flow-matching \cite{zhangFlowpolicyEnablingFast2025}, have successfully reduced inference complexity by collapsing or skipping denoising steps. However, \emph{training} still requires sampling all noise levels thoroughly, posing a significant challenge for generalist agents. Such agents need to efficiently scale to diverse tasks, instructions, and observation modalities.

Compared to task-specific diffusion policies, generalist agents typically need much larger models and training datasets with much more training steps, increasing training costs considerably. This challenge has been clearly shown in recent works such as Pi-Zero \cite{black$p_0$VisionlanguageactionFlow2024}, and 3D Diffuser Actor \cite{ke3DDiffuserActor2024}, where training can take multiple days on clusters with multiple GPUs—similar to general-purpose image generators like Stable Diffusion \cite{rombachHighresolutionImageSynthesis2022}.

\begin{figure}
    \centering
    \includegraphics[width=0.8\linewidth]{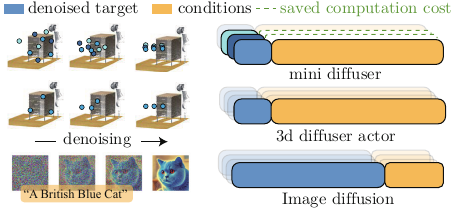}
    \caption{\textbf{Difference between image diffusion (bottom), state-of-art action diffusion \cite{ke3DDiffuserActor2024} (middle) and our mini-diffuser (top).} {A semantically meaningful image is denoised from fully random pixels, while structured and meaningful actions are denoised from random samples.} At token level, the denoised target (image) dominates token space in image diffusion (bottom row). By contrast, in action diffusion (middle and top rows), the denoised target (action) lies in a low-dimensional vector space relative to the conditioning inputs. By re-using the same condition for multiple action samples, Mini-diffuser can achieve a per-sample computation and memory cost that is significantly lower than 3D diffuser actor. }
    \label{fig:compare}
\end{figure}
We identify a critical but often overlooked asymmetry between robotic policy learning and image generation. In image generation, the condition (e.g., a text prompt) is typically smaller and simpler than the output (high-dimensional pixels). In contrast, robotic action generation usually has conditions (rich multimodal robot states including visual features, proprioception, and language instructions) that are much larger and more complex than the relatively low-dimensional action outputs.

This imbalance offers a unique opportunity to improve training efficiency. Specifically, during training, conditions remain the same across multiple noise-level predictions within a given context. Leveraging this, we propose Level-2 batching, a novel yet simple sampling strategy that reuses the same condition across multiple noise-level predictions, to enhance sample efficiency significantly. However, applying this strategy directly would cause redundant computations, as traditional network architectures would repeatedly process the same condition for each prediction.

To address this, we introduce a non-invasive mini-diffuser architecture, which employs masked global attention, sample-wise adaptive normalization, and local kernel-based feature fusion. These carefully selected modules avoid inter-sample dependencies, enabling the processing of large flattened Level-2 batches without additional memory usage or computational overhead. Consequently, we significantly scale effective batch sizes and reduce the number of gradient updates necessary for convergence.


In summary, our contributions are:

\begin{itemize}
    \item \textbf{Level-2 Batch Sampling for Condition-Element Asymmetry:} We formalize the asymmetry in diffusion-based policy training and introduce a two-Leveled batching method that significantly speeds up training by exploiting shared conditions.
    \item \textbf{Non-invasive Mini-Diffuser Architecture:} We design a compact diffusion policy architecture composed of non-invasive, condition-invariant layers, enabling efficient processing of large flattened Level-2 batches. This approach maintains most of the expressiveness of full-scale 3D diffusion policies while dramatically reducing training time and computational resources.
\end{itemize}

\begin{figure}
    \centering
    \includegraphics[width=0.8\linewidth]{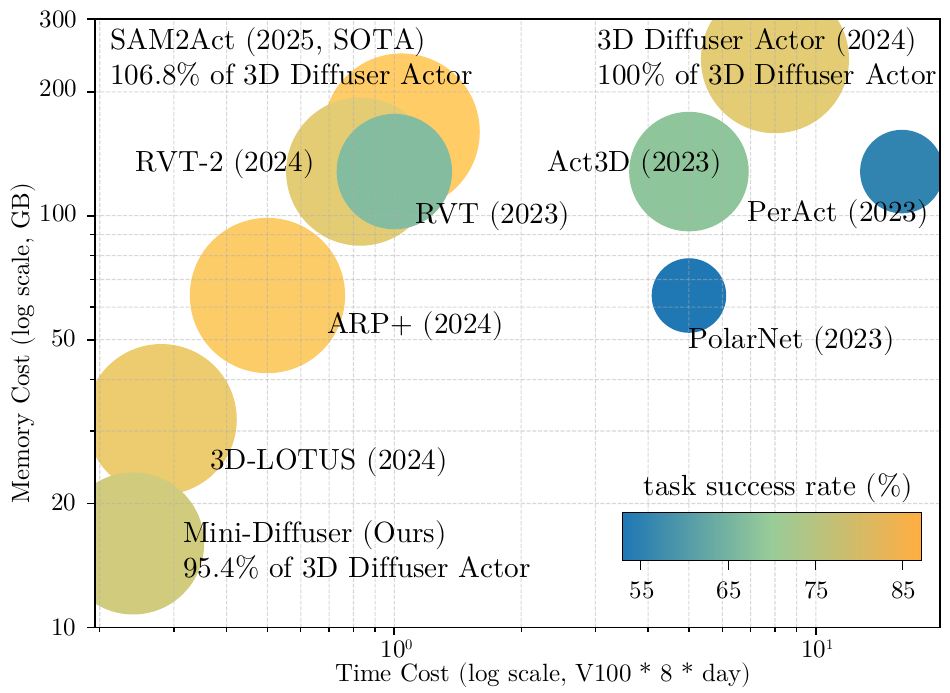}
    \caption{\textbf{Comparison with state of the arts in RLbench Peract-18 benchmark.} Our method by far takes least time and memory to train, while maintain 95\% of the performance of currently SOTA diffusion baed model.}
    \label{fig:bubble}
\end{figure}

Thanks to these improvements, we achieve by far the lowest training cost for high-capacity multitask diffusion policies while sacrificing only about 5\% of performance compared with current SOTA\cite{ke3DDiffuserActor2024}. Training efficiency comparisons using a unified time standard are highlighted in Figure \ref{fig:bubble}. Notably, our model can be trained end-to-end on a consumer-level GPU, such as an RTX 4090, in just 13 hours—while existing diffusion and non-diffusion methods typically require multiple GPUs and days of training. Real-world experiments further confirm that our approach preserves the robust multimodal action-generation capabilities that diffusion models are known for, ensuring reliable performance across diverse perceptual inputs.

\section{Related Works}

\subsection{Robot Learning from demonstration}

Earlier works on learning from demonstrations train deterministic policies with behavior cloning \cite{pomerleauAlvinnAutonomousLand1988}, mapping observations directly to actions. To better capture multimodal action distributions, later approaches discretize the action space and apply cross-entropy losses \cite{zengTransporterNetworksRearranging2021, chenPolarNet3DPoint2023, garciaGeneralizableVisionlanguageRobotic2024}, or leverage generative models such as VAEs \cite{mandlekarIrisImplicitReinforcement2020}, and diffusion models \cite{chiDiffusionPolicyVisuomotor2024, ze3DDiffusionPolicy2024}. Autoregressive training \cite{zhangAutoregressiveActionSequence2025, zhaoLearningFinegrainedBimanual2023a} and pretrained foundition models \cite{fangSAM2ActIntegratingVisual2025a} are also used for better capture spaitial and temporal features.

The other line of work is trying to broaden a single model's capability. Multi-task policies aim to generalize across variations of the same set if tasks or appearance changes in the environment. By incorporating multi-view perception together with language instructions. C2F-ARM \cite{jamesCoarsetofineQattentionEfficient2022a} and PerAct \cite{shridharPerceiveractorMultitaskTransformer2023} voxelize the workspace to localize target keyposes, while Act3D \cite{gervetAct3D3DFeature2023a} avoids voxelization by sampling 3D points and applying cross-attention to physical scene features. Robotic View Transformer (RVT) series \cite{goyalRvtRoboticView2023a, goyalRVT2LearningPrecise2024a} further improves spatial reasoning by projecting RGB-D inputs into multiple views and lifting them into 3D. To even go beyond training task sets and environment, Generalist policies such as RT-X series \cite{brohanRT1RoboticsTransformer2023, brohanRT2VisionlanguageactionModels2023, oneillOpenXembodimentRobotic2024}, Octo \cite{teamOctoOpensourceGeneralist2024} and OpenVLA \cite{kimOpenVLAOpensourceVisionlanguageaction2024a} adopt large transformer-based architectures to directly predict low-level actions from raw visual input. These models show strong scalability and task coverage, but often rely on massive datasets and train time to implicitly learn state and action representation together. While a single task policy can be trained within hours \cite{chenPolarNet3DPoint2023, chiDiffusionPolicyVisuomotor2024}, multi-task policies normally take days \cite{shridharPerceiveractorMultitaskTransformer2023}, and generalist policies take weeks on a cluster of cards \cite{kimOpenVLAOpensourceVisionlanguageaction2024a, black$p_0$VisionlanguageactionFlow2024}.

\subsection{Diffusion policies and their extensions}

 Recently, diffusion models have been increasingly explored in the field of embodied AI, showing promising progress in tasks that require nuanced decision-making and adaptive control. Early works \cite{ajayConditionalGenerativeModeling2022, reussGoalconditionedImitationLearning2023, chiDiffusionPolicyVisuomotor2024} demonstrated the effectiveness of diffusion models in low-dimensional control settings. Building on this, more recent efforts have extended diffusion-based approaches to complex 3D robotic manipulation tasks \cite{ze3DDiffusionPolicy2024, ke3DDiffuserActor2024}, achieving performance that surpasses traditional architectures.

Despite their success, applying diffusion models in 3D robotic domains presents significant challenges. These tasks involve intricate spatial representations and demand high-frequency decision-making, which conflicts with the inherently iterative and computationally intensive nature of diffusion-based training and denoising processes. Several methods propose skipping inference steps via hierarchical sampling \cite{xianChaineddiffuserUnifyingTrajectory2023a, maHierarchicalDiffusionPolicy2024}, or try replacing diffusion models with Consistency Models \cite{chenBoostingContinuousControl2024}, \cite{zhangFlowpolicyEnablingFast2025}. Though these method mitigate inference time efficiency. The training cost keeps high, especially in the multi-task training setting.


\section{Diffusion Model Formulation and Training}

\subsection{Problem Definition}

A multi-task robotic manipulation policy aims to predict an action vector $\boldsymbol{a}$ conditioned on the current state $\boldsymbol{s}$. To train such a policy, we use expert demonstrations in the form of temporally ordered state-action sequences $\{(\boldsymbol{s}_0, \boldsymbol{a}_0), \ldots, (\boldsymbol{s}_t, \boldsymbol{a}_t)\}$, consistent with prior work in multimodal imitation learning.

Each state $\boldsymbol{s}$ consists of a combination of modalities, including RGB-D images with known camera poses, proprioceptive signals such as joint angles and end-effector velocities and task-specific language instruction. These components may be sampled from a single timestep or a short history temporal window.

Each action $\boldsymbol{a}$ defines a low-level end-effector command or a short sequence of future commands. It is represented as a tuple:
\begin{equation}
\boldsymbol{a} = (\boldsymbol{a}_{\text{pos}}, \boldsymbol{a}_{\text{rot}}, a_{\text{open}}) \in \mathbb{R}^3 \times \mathbb{SO}(3) \times \{0, 1\},
\end{equation}
where $\boldsymbol{a}_{\text{pos}}$ is the 3D position, $\boldsymbol{a}_{\text{rot}}$ is the 3D rotation, and $a_{\text{open}}$ is the gripper open/close flag.

\subsection{Conditional Diffusion Model Formulation}

For simplicity and generality, we omit the real-world time index $t$ of $\boldsymbol{a}$ and $\boldsymbol{s}$ to avoid confusion with the denoising step index $k$ used in diffusion.

We aim to model the conditional probability distribution $p(\boldsymbol{a}|\boldsymbol{s})$ via a diffusion model. Given action-state pairs $(\boldsymbol{a}_0, \boldsymbol{s}) \sim q(\boldsymbol{a}, \boldsymbol{s})$, the forward diffusion process is defined as:
\begin{equation}
q(\boldsymbol{a}_k|\boldsymbol{a}_0, \boldsymbol{s}) = 
\mathcal{N}(\boldsymbol{a}_k;\sqrt{\bar{\alpha}_k} \boldsymbol{a}_0, (1 - \bar{\alpha}_k)\boldsymbol{I}),
\end{equation}
where $k \in \{1, \ldots, K\}$, $\bar{\alpha}_k = \prod_{j=1}^{k} (1 - \beta_j)$, and $\{\beta_j\}$ is a noise schedule defined by a pre-specified function [1] with their correspond terms $\sigma_j$ used for denoising. Although $\boldsymbol{s}$ does not affect the forward process directly, we include it for clarity, since our goal is to learn the conditional distribution $p(\boldsymbol{a}|\boldsymbol{s})$. As $k \rightarrow K$, the distribution $q(\boldsymbol{a}_K|\boldsymbol{a}_0)$ approaches a standard Gaussian $\mathcal{N}(0, \mathbf{I})$, ensuring that we can begin the reverse generation from pure white noise, and progressively apply the reverse diffusion steps conditioned on $\boldsymbol{s}$:
\begin{equation}
\boldsymbol{a}_{k-1} = \frac{1}{\sqrt{\alpha_k}} \left(\boldsymbol{a}_k - \frac{1 - \alpha_k}{\sqrt{1 - \bar{\alpha}_k}} \epsilon_\theta(\boldsymbol{a}_k, k, \boldsymbol{s})\right) + \sigma_k z,
\end{equation}
where $z \sim \mathcal{N}(0, \mathbf{I})$, $k = K, ..., 1$, and $\epsilon_\theta(\cdot)$ is a neural network parameterized by $\theta$. This iterative reverse process yields the final predicted action $\boldsymbol{a}_0$ conditioned on the state $\boldsymbol{s}$.

We train our model by minimizing the conditional denoising objective:
\begin{equation}
L(\theta) = \mathbb{E}_{(\boldsymbol{a}_0, \boldsymbol{s}), k, \epsilon} \left[ \| \epsilon - \epsilon_\theta(\boldsymbol{a}_k, k, \boldsymbol{s}) \|^2 \right].
\end{equation}

This objective teaches the model to predict the noise $\epsilon$ that was added to $\boldsymbol{a}_0$ during the forward process, thereby enabling accurate recovery of $\boldsymbol{a}_0$ during the reverse process.

In practice, training is performed over mini-batches. For a mini-batch of $N$ samples $\{(\boldsymbol{a}_0^{(i)}, \boldsymbol{s}^{(i)}, k^{(i)}, \epsilon^{(i)})\}_{i=1}^N$, the empirical loss becomes:
\begin{equation}
L(\theta) = \frac{1}{N} \sum_{i=1}^N \left\| \epsilon^{(i)} - \epsilon_\theta(\boldsymbol{a}_k^{(i)}, k^{(i)}, \boldsymbol{s}^{(i)}) \right\|^2.
\end{equation}

\subsection{Level-2 Mini-batch Sampling}

As mentioned earlier, a unique characteristic of robotic policy learning is the discrepancy in dimensionality between actions and states: $\text{dim}(\boldsymbol{a}) \ll \text{dim}(\boldsymbol{s})$. This motivates a specialized sampling strategy, which we call \emph{Level-2 batching}, where multiple noise-level predictions are computed under shared state conditions. Our mini-batch is organized into two Levels: In \textbf{Level-1}, we sample $B$ state-action pairs independently:
\begin{equation}
\{(\boldsymbol{s}_i, \boldsymbol{a}_0^{(i)})\}_{i=1}^{B}, \quad (\boldsymbol{s}_i, \boldsymbol{a}_0^{(i)}) \sim q(\boldsymbol{a}, \boldsymbol{s}).
\end{equation}
In \textbf{Level-2}, for each Level-1 pair, we independently sample $M$ noise-timestep pairs:
\begin{equation}
\{(k^{(j)}, \epsilon^{(j)})\}_{j=1}^{M},\ k^{(j)} \sim U(1, K),\ \epsilon^{(j)} \sim \mathcal{N}(0, \mathbf{I}).
\end{equation}
We finally flatten the samples to a single batch of size $B \cdot M$:
\begin{equation}
\label{eq:Level12}
\{(\boldsymbol{a}_k^{(n)}, k^{(n)}, \boldsymbol{s}^{(n)})\}_{n=1}^{B \cdot M}, \quad n = (i - 1)M + j,
\end{equation}
where
\begin{equation}
\boldsymbol{a}_k^{(n)} = \sqrt{\bar{\alpha}_{k^{(n)}}} \boldsymbol{a}_0^{(i)} + \sqrt{1 - \bar{\alpha}_{k^{(n)}}} \epsilon^{(n)}, \quad \boldsymbol{s}^{(n)} = \boldsymbol{s}^{(i)}.
\end{equation}
The final training loss becomes:
\begin{equation}
L(\theta) = \frac{1}{B \cdot M} \sum_{i=1}^{B} \sum_{j=1}^{M} \left\| \epsilon^{(j)} - \epsilon_\theta(\boldsymbol{a}_k^{(j)}, k^{(j)}, \boldsymbol{s}^{(i)}) \right\|^2,
\end{equation}
where all Level-2 samples within the same Level-1 batch share the same state condition $\boldsymbol{s}$. The key objective is to approximate the learning effect of having $N=B \cdot M$ independently sampled pairs from $q(\boldsymbol{a}, \boldsymbol{s})$, while incurring only the cost of processing $B$ unique state conditions. In the next section, we describe the network architecture designed to support this efficient reuse of condition encoding.

\section{Model Architecture Design}

To support Level-2 batching, our model processes all noised action samples and shared condition information in a single forward pass. After project them into feature space of dimension $d$, we concatenate per-sample action tokens and condition tokens into a flattened sequence:
\begin{equation}
\boldsymbol{Z} = [\boldsymbol{z}^{(1)}, \boldsymbol{z}^{(2)}, \ldots, \boldsymbol{z}^{(M)}, \boldsymbol{h}^{(\text{share})}_{\text{vis}}, \boldsymbol{h}^{(\text{share})}_{\text{ctx}}].
\end{equation}
Each $\boldsymbol{z}^{(m)} \in \mathbb{R}^{L \times d}$ is the token sequence of the $m$-th noised action sample, with a sequence length of $L$. The shared visual condition tokens $\boldsymbol{h}^{(\text{share})}_{\text{vis}} \in \mathbb{R}^{N \times d}$ are projected from RGB-D pixels lifted into 3D space, and $\boldsymbol{h}^{(\text{share})}_{\text{ctx}} \in \mathbb{R}^{C \times d}$ represent non-spatial features such as language and proprioception.

We use a Transformer-style architecture in which the entire sequence $\boldsymbol{Z}$ is linearly projected into queries $\boldsymbol{Q}$, keys $\boldsymbol{K}$, and values $\boldsymbol{V}$. For spatial tokens (actions and 3D visual points), we apply 3D rotary positional encoding (RoPE)\cite{suRoFormerEnhancedTransformer2024}\cite{gervetAct3D3DFeature2023a} to capture relative spatial relationships. For context tokens, we add a learned modality-specific embedding.

\label{rd:vanilla}A standard Transformer architecture uses multi-layer self-attention, which enables global information sharing but introduces a risk of information leakage across independently sampled action sequences. This is especially problematic under Level-2 batching, where each sample must remain isolated. To address this, we replace standard attention layers with specialized non-invasive modules that allow efficient condition querying while preserving isolation between noised samples.

\begin{figure*}[t]
    \centering
    \includegraphics[width=1\linewidth]{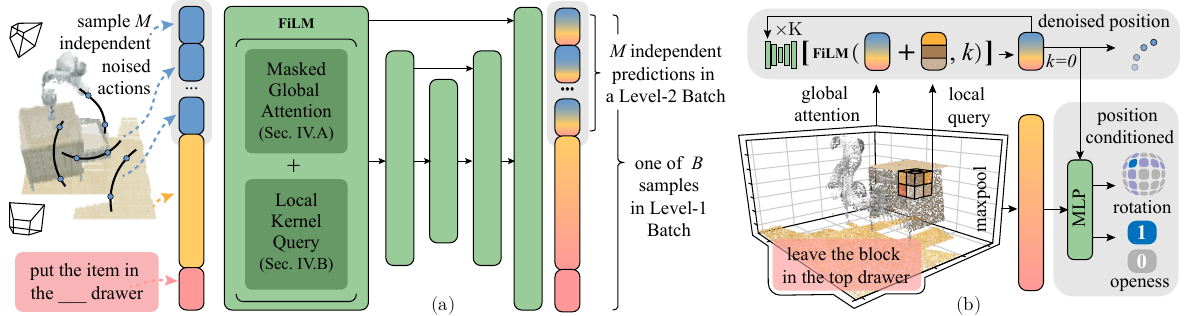}
    \caption{\textbf{Mini-diffuser model structure.} (a) During training phase, $B$ samples of the states form a Level-1 batch, where $M$ noise actions are sampled independently under the same state conditions, building a Level-2 batch. Tokens are flattened and feed into a multi-layer model contains Masked attention module, local query module, and FiLM layers. (b) During inference phase, denoising is applied only to the end-effector position. Rotation and gripper state are predicted separately via classification heads conditioned on the final denoised position.}
    \label{fig:pipeline}
\end{figure*}

\subsection{Masked Global Attention}

We apply a combination of self- and cross-attention using masked attention, which enables selective communication and avoids information leakage. The masked attention is defined as:
\begin{align}
\boldsymbol{Z}' &= \text{MaskedAttn}(\boldsymbol{Q}, \boldsymbol{K}, \boldsymbol{V}) \notag \\
&= \text{Softmax}\bigg( \frac{\boldsymbol{Q}\boldsymbol{K}^\top 
- \text{Inf} \cdot (1 - \boldsymbol{M}_{\text{QK}})}{\sqrt{d}} \bigg)\boldsymbol{V},
\end{align}
where $\boldsymbol{M}_{\text{QK}}$ is a binary mask matrix that defines which tokens may attend to which others, and $\text{Inf}$ is a large constant used to suppress masked entries in the attention logits.

The masking matrix $\boldsymbol{M}_{\text{QK}}$ is constructed to satisfy: (i) An action sample attends to itself and shared conditions, but not to other action samples, and (ii) shared conditions do not attend back to action samples. The structure of the mask is:
\begin{equation}
\boldsymbol{M}_{\text{QK}} =
\begin{bmatrix}
\boldsymbol{I}_{M \times M} \otimes \boldsymbol{J}_{L \times L} & \boldsymbol{J}_{(M \cdot L) \times (N + C)} \\
\boldsymbol{0}_{(N + C) \times (M \cdot L)} & \boldsymbol{J}_{(N + C) \times (N + C)}
\end{bmatrix},
\end{equation}
where $\boldsymbol{I}_{M \times M} \otimes \boldsymbol{J}_{L \times L}$ constructs $M$ diagonal intra-sample blocks of size $L \times L$ using the Kronecker product $\otimes$. The top-right block enables all action tokens to perform cross-attention to condition tokens, while the bottom-left zero block prevents condition tokens from attending back to samples.

This masked attention mechanism is central to enabling efficient Level-2 batching, allowing all samples to share condition processing while maintaining proper sample-wise independence during training.

\subsection{Local Kernel-Based Query}\label{sec:local}

While masked attention captures global structure, we further enhance spatial grounding through a local feature aggregation module that strengthens the influence of nearby 3D geometry. This is particularly important for guiding the end-effector toward target regions based on local scene structure.

We discretize 3D space into voxel bins using an \emph{octree}-like subdivision. Each point $\boldsymbol{p}_i = (x_i, y_i, z_i)$ is mapped to a bin index $(x_{\text{bin}}, y_{\text{bin}}, z_{\text{bin}})$ by downsampling the coordinates (in this paper, dividing by $2^4$). These indices define a voxel grid of size $X \times Y \times Z$. Each bin is then hashed into a unique index:
\begin{equation}
\phi(\boldsymbol{p}_i) = x_{\text{bin}} \cdot (Y \cdot Z) + y_{\text{bin}} \cdot Z + z_{\text{bin}}.
\end{equation}
Given a query position $\boldsymbol{q}_j = (x_q, y_q, z_q)$, we define a local neighborhood as the surrounding $3 \times 3 \times 3$ voxel block. For each offset $(\delta x, \delta y, \delta z) \in \{-1, 0, 1\}^3$, we compute the hash of the neighboring bin:
\begin{equation}
\phi_{\text{nbr}} = \phi(x_q + \delta x, y_q + \delta y, z_q + \delta z).
\end{equation}
Let $\mathcal{P}(\phi_{\text{nbr}})$ denote the set of points that fall into the bin with hash $\phi_{\text{nbr}}$. If the bin is non-empty, we average their features:
\begin{equation}
\bar{\boldsymbol{h}}(\phi_{\text{nbr}}) = \frac{1}{|\mathcal{P}(\phi_{\text{nbr}})|} 
\sum_{i \in \mathcal{P}(\phi_{\text{nbr}})} \boldsymbol{h}_i.
\end{equation}
If no points fall into the bin, we define $\bar{\boldsymbol{h}}(\phi_{\text{nbr}}) = \boldsymbol{0}$.

We associate each relative offset with a learnable weight matrix 
$\boldsymbol{W}_{\delta x, \delta y, \delta z} \in \mathbb{R}^{d \times d}$ {to do feature projection without changing of hidden dimension $d$}, which together form a $3 \times 3 \times 3$ convolutional kernel in 3D space. Unlike standard convolutions applied over dense grids, this kernel is only applied at the query position $\boldsymbol{q}_j$ to gather context from its spatial neighborhood:
\begin{equation}
\boldsymbol{h}_{\text{local}}(\boldsymbol{q}_j) = 
\sum_{\delta x, \delta y, \delta z} 
\boldsymbol{W}_{\delta x, \delta y, \delta z} \cdot 
\bar{\boldsymbol{h}}(\phi_{\text{nbr}}).
\end{equation}
This operation is non-invasive: it retrieves spatial context from the environment without modifying shared condition features, making it fully compatible with Level-2 batching. During training, each noised sample gathers local cues around its position. During inference, these query locations gradually shift toward the target end-effector position as the denoising process progresses.

\subsection{Per-Sample Modulation Conditioned on Noise Step}

To allow the model to adapt to different stages of the denoising process, we apply per-sample modulation using Feature-wise Linear Modulation (FiLM) \cite{perezFilmVisualReasoning2018}. For each sample at a diffusion timestep $k$, we embed the timestep index and transform it into a pair of scale $\boldsymbol{\gamma}$ and shift vectors $\boldsymbol{\beta}$ using a lightweight MLP. These vectors are then applied to intermediate features within the network using affine transformation. Formally, the FiLM layer modulates a feature vector $\boldsymbol{z}$ as:
\begin{equation}
\text{FiLM}(\boldsymbol{z}; \boldsymbol{\gamma}, \boldsymbol{\beta}) = 
\boldsymbol{\gamma} \odot \boldsymbol{z} + \boldsymbol{\beta}.
\end{equation}
This simple yet effective mechanism allows the network to dynamically adjust its behavior for different noise levels—handling coarse predictions at early timesteps and refining details at later ones. Crucially, FiLM is applied independently to each sample, ensuring that no information is shared across the Level-2 batch.

\subsection{Design Choices}

We explore two variants that integrate our three non‐invasive building blocks—masked attention, local kernel–based query, and FiLM modulation—into transformer‐style diffusion models. The first variant serves as a lightweight modified baseline which we used for ablation, while the second, which we refer to as the \emph{Mini-Diffuser}, represents our fully optimized architecture.
\subsubsection{Minimal Modifications to 3D Diffuse Actor}
In the simplest setup, we replace the self-attention modules in each transformer layer of 3D Diffuse Actor \cite{ke3DDiffuserActor2024} with our non-invasive counterparts. This drop-in replacement preserves the original layer-wise architecture and requires no additional structural changes. Importantly, this modification alone enables Level-2 batching during training, allowing us to isolate and evaluate the resulting memory and computational savings. We demonstrate the effectiveness of this baseline in Sec.~\ref{sec:design}. 

\subsubsection{Mini-Diffuser}\label{rd:unet}
To further improve parameter training efficiency, we adopt the U-Net-style architecture of Point Transformer v3 (PTv3)~\cite{wuPointTransformerV32024a}. PTv3 uses transformer layers within a U-Net framework~\cite{ronnebergerUnetConvolutionalNetworks2015}, combining downsampling and upsampling stages to compute compact 3D-aware latent features. This hierarchical structure reduces memory usage and computation in the deeper middle layers. Additionally, we can cache the point indices used during down-sampling, allowing us to accelerate the local kernel query among point neighborhoods described in Sec.~\ref{sec:local}. Despite inheriting the PTv3 backbone, we replace all internal transformer blocks with non-invasive counterparts. As a result, we cannot directly use pretrained PTv3 weights and the training time saving does not come from pretrained weights.

We also adopt a decoupled action head. Since 3D RoPE applies only to spatial position coordinates, end-effector rotation and gripper states are not spatially aligned with the point-based token structure and may introduce noise if fused too early. Unlike 3D Diffuser Actor, which embeds all action components into a single denoised token, we adopt a decoupled design following prior works~\cite{goyalRVT2LearningPrecise2024a, ajayConditionalGenerativeModeling2022}: Denoising is applied only to the end-effector position. Rotation and gripper state are predicted separately via classification heads (with cross-entropy loss), conditioned on the final denoised position. This design preserves the multimodal nature of action generation, as the model can flexibly associate different discrete rotations or gripper commands with the different predicted position.

\section{Experiments}

We evaluate our Mini-Diffuser for multi-task robotic manipulation from demonstrations in both simulation and real-world settings. Our primary simulated benchmark is RLBench~\cite{jamesRlbenchRobotLearning2020}, a widely adopted platform for vision-language manipulation tasks. Our experiments aim to answer the following questions:

(1) How does Mini-Diffuser perform compared to state-of-the-art methods?

(2) How do our proposed architectural design choices contribute to training acceleration and sample efficiency?

(3) Can Mini-Diffuser maintain competitive task performance despite significantly reduced training time and resources?

\subsection{Simulation Benchmark}

\subsubsection{Datasets}
We evaluate Mini-Diffuser on the multi-task RLBench benchmark proposed by PerAct~\cite{shridharPerceiveractorMultitaskTransformer2023}, consisting of 18 tasks and 249 task variations. These tasks require generalization across diverse goal configurations including object types, colors, shapes, and spatial arrangements. Each method is trained with 100 demonstrations per task, which include multi-view RGB-D images, language goals, and extracted end-effector keyposes. Following prior work~\cite{shridharPerceiveractorMultitaskTransformer2023, goyalRvtRoboticView2023a, ke3DDiffuserActor2024}, we segment trajectories into keyposes and only predict the next keypose at each time step. For evaluation, each method is tested across 300 unseen episodes per task, using three different random seeds.

\subsubsection{Baselines}
We compare Mini-Diffuser with the following state-of-the-art baselines:

\begin{itemize}
    \item \textbf{PerAct}~\cite{shridharPerceiveractorMultitaskTransformer2023}: A voxel-based policy that applies global self-attention over a 3D voxel grid.
    \item \textbf{Act3D}~\cite{gervetAct3D3DFeature2023a}: A point cloud-based approach using coarse-to-fine refinement for action prediction.
    \item \textbf{RVT} \cite{goyalRvtRoboticView2023a} / \textbf{RVT-2}~\cite{goyalRVT2LearningPrecise2024a}: Policies using rendered multi-view images for policy learning. RVT-2 adds a hierarchical action selection mechanism.
    \item \textbf{3D Diffuse Actor}~\cite{ke3DDiffuserActor2024}: The current state-of-the-art diffusion-based policy on RLBench.
    \item \textbf{SAM2Act}~\cite{fangSAM2ActIntegratingVisual2025a}: The strongest baseline on this benchmark, integrating the Segment Anything Model (SAM) \cite{kirillovSegmentAnything2023} to enhance visual perception.
\end{itemize}

\subsubsection{Results}

\begin{table}[h]
\caption{Summary of Table 2 - metrics and reported hardware for Multi-task RLBench. \\ }

\begin{tabular}{lcccc}
\toprule
Method & Avg.\ Suc.\ (\%) & Norm.\ Time & Memory (GB) & Reported Hardware \\
\midrule
PerAct           & 49.4               & 128         & 128          & V100$\times$8$\times$16 days \\
RVT              & 62.9               & 8           & 128          & V100$\times$8$\times$1 day \\
Act3D            & 63.2               & 40          & 128          & V100$\times$8$\times$5 days \\
RVT-2            & 81.4               & 6.6         & 128          & V100$\times$8$\times$20 hours \\
3D-Dif-Actor     & 81.3 \textcolor{gray}{(100\%)}       & 39 \textcolor{gray}{(100\%)}  & 240 \textcolor{gray}{(100\%)}  & A100$\times$6$\times$6 days \\
SAM2Act          & 86.8               & 8.3         & 160          & H100$\times$8$\times$12 hours \\
Mini-diffuser    & 77.6 \textcolor{gray}{(95.4\%)}      & \textbf{1.9 \textcolor{gray}{(4.8\%)}} & \textbf{16 \textcolor{gray}{(6.6\%)}} & 4090$\times$13 hours or A100$\times$1 day \\
\bottomrule
\end{tabular}
\label{table:summary_metrics}
\end{table}

\begin{table*}[h]
\centering
\caption{Multi-task RLBench benchmark results. Metrics include task success rate (\%), Normalized training time (V100*8*days), and peak memory usage (GB). Mini-Diffuser uses a single RTX 4090 GPU and achieves 95\% of 3D Diffuse Actor's success rate with minimal compute.\\}
\resizebox{\textwidth}{!}{%
\setlength{\tabcolsep}{3pt}
\begin{tabular}{@{}l|ccc|ccccccccc@{}}
\toprule
\multicolumn{1}{c|}{Method} & Avg.\ Suc.\ (\%) & Norm. Time & Memory (GB) & Close Jar & Drag Stick & Insert Peg & Meat Grill & Open Drawer & Place Cups & Place Wine & Push Buttons & Put in Cup \\ \midrule
PerAct  & 49.4 & 128 & 128 & 55.2 ± 4.7 & 89.6 ± 4.1 & 5.6 ± 4.1 & 70.4 ± 2.0 & 88.0 ± 5.7 & 2.4 ± 3.2 & 44.8 ± 7.8 & 92.8 ± 3.0 & 28.0 ± 4.4 \\
RVT & 62.9 & 8 & 128 & 52.0 ± 2.5 & 92.2 ± 1.6 & 11.0 ± 4.0 & 88.0 ± 2.5 & 71.2 ± 6.9 & 4.0 ± 2.5 & 91.0 ± 5.2 & 100.0 ± 0.0 & 49.6 ± 3.2 \\
Act3D & 63.2 & 40 & 128 & 96.8 ± 3.2 & 80.8 ± 6.4 & 24.0 ± 8.4 & 95.2 ± 1.6 & 78.4 ± 11.2 & 3.2 ± 3.2 & 59.2 ± 9.8 & 92.8 ± 3.0 & 67.2 ± 3.9 \\
RVT-2  & 81.4 & 6.6 & 128 & 100.0 ± 3.6 & 97.2 ± 1.6 & 4.2 ± 1.2 & 99.0 ± 1.7 & 74.0 ± 6.9 & 14.0 ± 2.8 & 95.0 ± 3.2 & 100.0 ± 0.0 & 66.0 ± 4.5 \\
3D-Dif-Actor & 81.3\textcolor{gray}{(100\%)} & 39\textcolor{gray}{(100\%)} & 240\textcolor{gray}{(100\%)} & 96.0 ± 2.5 & 100.0 ± 0.0 & 65.6 ± 4.1 & 96.8 ± 1.6 & 89.6 ± 4.1 & 24.0 ± 7.6 & 93.6 ± 4.8 & 98.4 ± 2.0 & 85.0 ± 4.1 \\
SAM2Act & 86.8 & 8.3 & 160 & 99.0 ± 2.0 & 99.0 ± 2.0 & 84.0 ± 5.7 & 98.0 ± 2.3 & 83.0 ± 6.0 & 47.0 ± 6.0 & 93.0 ± 3.8 & 100.0 ± 0.0 & 75.0 ± 3.8 \\
Mini-diffuser & 77.6\textcolor{gray}{(95.4\%)} & \textbf{1.9}\textcolor{gray}{(4.8\%)} & \textbf{16}\textcolor{gray}{(6.6\%)} & \textbf{98.7 ± 0.5} & 97.3 ± 0.5 & \textbf{68.0 ± 1.5} & \textbf{100.0 ± 0.0} & 85.3 ± 3.7 & 16.0 ± 1.6 & 93.3 ± 2.1 & \textbf{100.0 ± 0.0} & 73.3 ± 5.4 \\ \midrule
\multicolumn{1}{c|}{Method} & \multicolumn{3}{c|}{reported hardware} & Put in Drawer & Put in Safe & Screw Bulb & Slide Color & Sort Shape & Stack Blocks & Stack Cups & Sweep Dust & Turn Tap \\ \midrule
PerAct  & \multicolumn{3}{c|}{V100$\times8\times16\text{days}$} & 51.2 ± 4.7 & 84.0 ± 3.6 & 17.6 ± 2.0 & 74.0 ± 13.6 & 16.8 ± 4.7 & 26.4 ± 3.2 & 2.4 ± 2.0 & 52.0 ± 0.0 & 88.0 ± 4.4 \\
RVT  & \multicolumn{3}{c|}{V100$\times8\times1\text{day}$} & 88.0 ± 5.7 & 91.2 ± 3.0 & 48.0 ± 4.9 & 81.6 ± 2.8 & 36.0 ± 2.5 & 28.8 ± 3.9 & 26.4 ± 2.4 & 72.0 ± 0.0 & 93.6 ± 4.1 \\
Act3D  & \multicolumn{3}{c|}{V100$\times8\times5\text{days}$} & 91.2 ± 6.2 & 95.2 ± 4.0 & 32.8 ± 6.9 & 96.0 ± 2.5 & 29.6 ± 3.2 & 4.0 ± 3.6 & 6.3 ± 2.0 & 86.4 ± 6.5 & 94.4 ± 2.0 \\
RVT-2  & \multicolumn{3}{c|}{V100$\times8\times20\text{hours}$} & 92.0 ± 0.0 & 96.0 ± 1.8 & 88.0 ± 4.9 & 81.0 ± 4.8 & 35.0 ± 7.1 & 80.0 ± 2.8 & 69.0 ± 5.9 & 100.0 ± 0.0 & 99.0 ± 1.7 \\
3D-Dif-Act & \multicolumn{3}{c|}{A100$\times6\times6\text{days}$} & 96.0 ± 3.6 & 97.6 ± 2.0 & 82.4 ± 4.2 & 97.6 ± 3.2 & 44.0 ± 4.4 & 68.3 ± 3.3 & 47.2 ± 8.5 & 84.0 ± 4.4 & 99.2 ± 1.6 \\
SAM2Act & \multicolumn{3}{c|}{H100$\times8\times12\text{hours}$} & 99.0 ± 2.0 & 98.0 ± 2.3 & 89.0 ± 2.3 & 86.0 ± 4.0 & 64.0 ± 4.6 & 76.0 ± 8.6 & 78.0 ± 4.0 & 99.0 ± 2.0 & 96.0 ± 5.7 \\
Mini-diffuser & \multicolumn{3}{c|}{4090$\times13\text{hours}$ or A100$\times1\text{day}$} & 96.0 ± 4.8 & 94.7 ± 0.4 & 77.3 ± 3.7 & \textbf{98.7 ± 0.1} & 28.0 ± 6.0 & 38.7 ± 4.0 & \textbf{48.0 ± 1.1} & 94.7 ± 0.9 & 89.3 ± 1.0 \\ \bottomrule
\end{tabular}

}
\label{table:main_results-18}
\end{table*}

Table~\ref{table:main_results-18} summarizes the results across all tasks. Mini-Diffuser achieves a strong average success rate while drastically reducing computational overhead. Specifically, it reaches \textbf{95.6\%} of the average task performance of 3D Diffuse Actor using only \textbf{4.8\%} of its training time and \textbf{6.6\%} of its memory consumption.

Remarkably, Mini-Diffuser outperforms 3D Diffuse Actor on 5 tasks—highlighted in bold in Table~\ref{table:main_results-18}—despite a drastic reduction in training time, and remains competitive on 9 others. These results validate the effectiveness of our lightweight architectural design and demonstrate the benefits of efficient batch sampling in diffusion-based policy learning. In terms of hardware requirements, Mini-Diffuser can be trained end-to-end on a single RTX 4090 GPU in under 13 hours, or on a single A100 GPU in one day, whereas state-of-the-art baselines require multi-GPU clusters running for several days. This efficiency makes Mini-Diffuser a practical solution for rapid experimentation and real-world deployment.

\subsection{Ablation Study}

We assess the impact of Mini-Diffuser’s core components through ablation experiments, summarized in Table~\ref{table:abla}. Unless otherwise specified, all models are trained with a Level-1 batch size of $B=100$ and a Level-2 batch size of $M=64$, on a subset of RLBench: \emph{Stack Block}, \emph{Slide Color}, and \emph{Turn Tap}.

\subsubsection{Effect of Level-2 Batching.} The primary innovation of Mini-Diffuser is the Level-2 batching strategy, which increases effective sample coverage without proportional increases in memory or compute. We evaluate performance under varying $M$. When $M=1$, Level-2 batching is disabled and training defaults to conventional sampling.

\begin{figure}[h]
    \centering
    \includegraphics[width=0.8\linewidth]{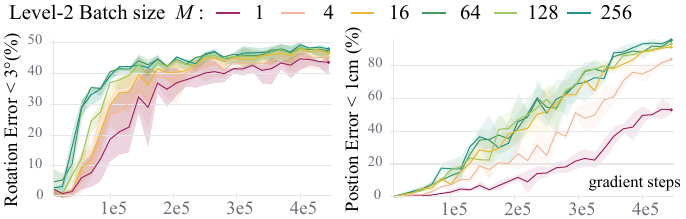}
    \caption{\textbf{Policy Learning Efficiency.} The y-axis shows the proportion of generated actions with errors below specified thresholds (e.g., $<1$cm or $<$3°), indicating successful diffusion. Increasing the Level-2 batch size accelerates convergence at the same number of gradient steps (x-axis).}

    \label{fig:t2batch128}
\end{figure}

Fig. \ref{fig:t2batch128} illustrates learning curves for different $M$ values. Larger $M$ accelerates convergence. This highlights that even without architectural changes, our Level-2 batching strategy alone yields substantial efficiency gains, though the benefit saturates beyond $M=128$. We attribute this to two factors: (i) Too large batches reduce gradient variance and diminish the stochasticity that benefits generalization; (ii) Level-2 batches are after all `fake' batches: they reuse the same condition across samples, limiting diversity relative to fully independent samples. 

Table~\ref{table:abla} further compares per-step memory and compute cost under different batching configurations. Increasing the Level-2 batch size $M$ to 64 results in 64 times more training samples being processed per step, yet introduces only a 3\% increase in memory usage and a 7\% increase in compute time. By contrast, achieving a similar total batch size through Level-1 is not possible. Only increasing $B$ by one time leads to nearly one time increase in both memory and computation as well — an expected result of scaling real batch size.

\subsubsection{Impact of Architectural Choices.}\label{sec:design} Table~\ref{table:abla} also evaluates other design decisions. When we revert Mini-Diffuser to match the original 3D Diffuse Actor—by replacing the PTv3 backbone and action head—performance remains comparable, but training time increases by 18\% and memory usage by 44\%.  Removing the 3D RoPE module leads to severe overfitting, decreasing success by 13.4\%, indicating the critical role of relative spatial encoding. Local kernel-based queries contribute a modest 1\% improvement in success rate, but we retain them as they help stabilize early training and add negligible computational cost.


\begin{table}[h]
\centering
\caption{Ablation on Duo-Level batches and Model Components.}
\begin{tabular}{@{}lccc@{}}
\toprule
 & \begin{tabular}[c]{@{}c@{}}consistent \\ memory cost\end{tabular} & \begin{tabular}[c]{@{}c@{}}iteration time\\ per decent\end{tabular} & \begin{tabular}[c]{@{}c@{}}Avg. Suc. \\ after 1e5\end{tabular} \\ \midrule
B=100 M=64 & 102.2\% & 106.3\% & 78.3 \\
B=100 M=1 & 100\% & 100\% & 44.1 \\
B=200 M=1 & 188.8\% & 176.6\% &  50.8 \\
\midrule
w.o. 3D ROPE & 101.2\% & 101.3\% & 67.8 \\
w.o. PTv3 backbone & 147.7\% & 125.5\% & 79.1 \\
w.o. local Conv & 102.2\% & 106.2\% & 77.9 \\ \bottomrule
\end{tabular}
\label{table:abla}
\end{table}

\subsection{Real-World Evaluation}

\begin{figure}[h]
    \centering
    \includegraphics[width=1\linewidth]{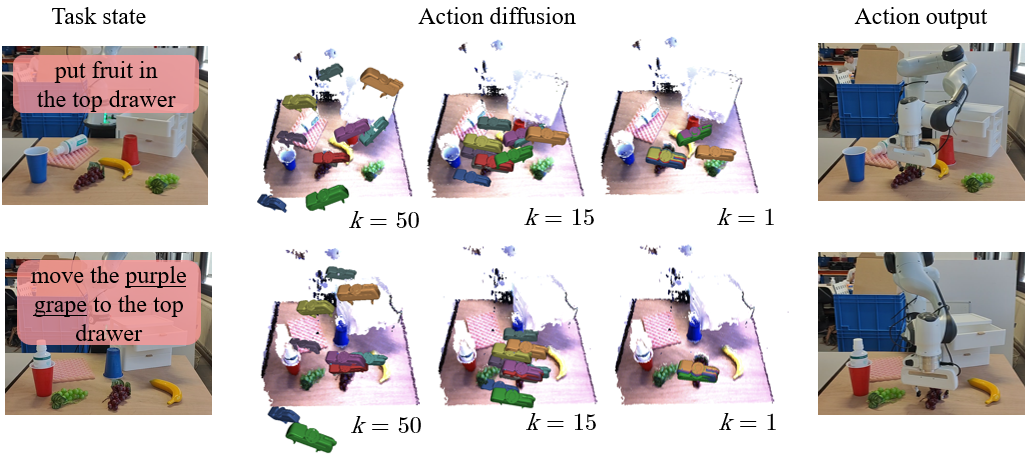}
    \caption{\textbf{Real World Mini-Diffuser.} We visualize 10 action candidates along the denoising trajectory, though only one is executed. Mini-Diffuser preserves the core strengths of diffusion-based actors. In the top row, when multiple actions are valid under the same instruction, Mini-Diffuser exhibits multi-modal behavior. In contrast, when language instructions differ but the visual scene remains unchanged, the model generates distinct actions that align precisely with the task description.}
    \label{fig:realworld}
\end{figure}

We validate Mini-Diffuser on 12 real-world manipulation tasks, repeating the same set of tasks used in 3D Diffuse Actor. We use a Franka Emika Panda robot equipped with a Roboception RGB-D camera mounted front-facing. RGB-D images are captured at 960 \texttimes{} 540 resolution and downsampled to a colored point cloud with no more than 4000 data points.

Each task is trained with 10 demonstrations collected by a human demonstrator. Demonstrations naturally include variation and multimodal behavior. For instance, in the "put fruit into drawer" task, different fruit and trajectories are used across demonstrations. In "insert peg", the user chooses one of multiple valid holes.

We evaluate 10 unseen trials per task and report success rates in Table~\ref{tab:realworld-results}. Mini-Diffuser achieves strong generalization and reproduces multimodal behavior effectively, conditioned on scene and language. Fig. \ref{fig:realworld} visualizes key state-action step under different instructions. Our model exhibits consistent grounding of language and spatial context, showing capabilities comparable to 3D Diffuse Actor.

\begin{table}[t]
\centering
\caption{Real-world success rates across 10 tasks (\%).}
\label{tab:realworld-results}
\setlength{\tabcolsep}{3pt}
\begin{tabular}{@{}cccccc@{}}
\toprule
Close Drawer & Put Mouse & Insert Peg & Put Grape & Fruit in \\ \midrule
100 & 100 & 50 & 70 & 80 \\ \midrule
Stack Block & Press Stapler & Sort Shape & Open Drawer & Close Box \\ \midrule
60 & 100 & 60 & 30 & 100 \\ \bottomrule
\end{tabular}
\end{table}

\section{Discussion and Conclusion}

Mini-Diffuser revisits diffusion-based policy learning with a focus on efficiency and practicality. Contrary to the common belief that diffusion models are slower than non-diffusion counterparts, our results show that with the right architectural design and batch sampling strategy, training time can be drastically reduced. While inference remains iterative, our architecture is compatible with step-skipping techniques like DDIM \cite{songDenoisingDiffusionImplicit2020} or Flow Matching \cite{zhangFlowpolicyEnablingFast2025}, which can further reduce runtime during deployment. Another potiential improvement can be addressing the limitations shared by most 3D-based manipulation policies, including reliance on camera calibration and depth input, and a focus only on quasi-static tasks, by extending Mini-Diffuser to dynamic settings with velocity control. 

Overall, Mini-Diffuser provides a fast, simple, and scalable baseline for multi-task manipulation. It can not only serves as a practical recipe for efficient policy training, but also has the potential to become a flexible platform for rapid experimentation and future research in architecture design, training strategies, and real-world robotic generalization.

\begin{ack}
This work is supported by Interne Fondsen KU Leuven/Internal Funds KU Leuven (C2E/24/034).
The resources and services used in this work were provided by the VSC (Flemish Supercomputer Center), funded by the Research Foundation - Flanders (FWO) and the Flemish Government. Part of calculations were also run on the Euler cluster of ETH Zürich
\end{ack}





\bibliographystyle{IEEEtran}
\bibliography{MiDi-RAL25}

\end{document}